\documentclass{article}

\usepackage{arxiv}

\usepackage[utf8]{inputenc} 
\usepackage[T1]{fontenc}    
\usepackage{hyperref}       
\usepackage{url}            
\usepackage{booktabs}       
\usepackage{nicefrac}       
\usepackage{microtype}      
\usepackage{lipsum}		
\usepackage{natbib}
\usepackage{doi}
\usepackage{amsmath,amssymb,amsfonts} 
\usepackage{algorithm,algorithmic}
\usepackage{subfig} 
\usepackage{url} 
\usepackage{textcomp}
\usepackage{xcolor}
\usepackage{siunitx}
\usepackage{enumitem}
\usepackage{etoolbox}
\usepackage{graphics} 
\usepackage{multirow}
\usepackage{epstopdf, epsfig} 
\graphicspath{{figs/}}

\newcommand*\rot{\rotatebox{90}}

\newcommand{\RA}[1]{\textcolor{black}{#1}}
\newcommand{\RB}[1]{\textcolor{black}{#1}}
\newcommand{\RC}[1]{\textcolor{black}{#1}}

\title{A Protection Method of Trained CNN Model with Secret Key from Unauthorized Access}

\date{} 					

\author{AprilPyone~MaungMaung\\
Tokyo Metropolitan University, Tokyo, Japan\\
	\And
	Hitoshi~Kiya \\
  Tokyo Metropolitan University, Tokyo, Japan\\
}

\renewcommand{\shorttitle}{}

\hypersetup{
pdftitle={A Protection Method of Trained CNN Model with Secret Key from Unauthorized Access},
pdfsubject={},
pdfauthor={},
pdfkeywords={},
}

\begin{document}
\maketitle

\begin{abstract}
In this paper, we propose a novel method for protecting convolutional neural network (CNN) models with a secret key set so that unauthorized users without the correct key set cannot access trained models. The method enables us to protect not only from copyright infringement but also the functionality of a model from unauthorized access without any noticeable overhead. We introduce three block-wise transformations with a secret key set to generate learnable transformed images: pixel shuffling, negative/positive transformation, and FFX encryption. Protected models are trained by using transformed images. The results of experiments with the CIFAR and ImageNet datasets show that the performance of a protected model was close to that of non-protected models when the key set was correct, while the accuracy severely dropped when an incorrect key set was given. The protected model was also demonstrated to be robust against various attacks. Compared with the state-of-the-art model protection with passports, the proposed method does not have any additional layers in the network, and therefore, there is no overhead during training and inference processes.
\end{abstract}


\section{Introduction}
Convolutional neural networks (CNNs) are a type of deep neural network (DNN) inspired by the human visual system. Recent advances in deep learning show that CNNs have lead to major breakthroughs in computer vision~\cite{Lecun2015}. Impressively, the last ImageNet Large Scale Visual Recognition Challenge (ILSVRC) in 2017 proved that the image classification accuracy has surpassed the level of human performance (i.e., error rate of \SI{2.25}{\percent}). There is no doubt that CNNs have dominated visual recognition systems in many different applications.

However, training successful CNNs is very expensive because it requires a huge amount of data and fast computing resources (e.g., GPU-accelerated computing). For example, the ImageNet dataset contains about 1.2 million images, and training on such a dataset takes days and weeks even on GPU-accelerated machines. In fact, collecting images and labeling them will also consume a massive amount of resources. Moreover, algorithms used in training a CNN model may be patented or have restricted licenses. Therefore, trained CNNs have great business value. Considering the expenses necessary for the expertise, money, and time taken to train a CNN model, a model should be regarded as a kind of intellectual property. While distributing a trained model, an illegal party may also obtain a model and use it for its own service.

To protect the copyrights of trained models, researchers have adopted digital watermarking technology to embed watermarks into the models~\cite{2017-ICMR-Uchida, 2020-NCA-Le, 2019-NIPS-Fan, 2019-MIPR-Sakazawa, 2018-Arxiv-Rouhani, 2018-ACCCS-Zhang, 2018-Arxiv-Chen, 2018-USENIX-Yossi}. These works focus on identifying the ownership of a model in question. In reality, a stolen model can be directly used by an attacker without arousing suspicion. In addition, the stolen model can be exploited in many different ways such as through model inversion attacks~\cite{2015-CCCS-Fredrikson} and adversarial attacks~\cite{2014-ICLR-Szegedy}. To the best of our knowledge, the consequences of a stolen model have not been considered before in model protection research except for ownership verification. In this paper, we focus on protecting a model from misuse when it has been stolen by taking inspiration from an adversarial defense.

Recently, a key-based adversarial defense was proposed to combat adversarial examples~\cite{2020-ICIP-Maung,2020-Arxiv-Maung}, which was in turn inspired by perceptual image encryption methods, which were proposed for privacy-preserving machine learning~\cite{kawamura2020privacy} and encryption-then-compression systems~\cite{2018-ICCETW-Tanaka, 2019-Access-Warit, 2019-TIFS-Chuman, 2019-ICIP-Warit, 2019-APSIPAT-Warit, 2017-IEICE-Kurihara, chuman2017security}. The uniqueness of the key for the model in~\cite{2020-ICIP-Maung,2020-Arxiv-Maung} motivated us to use a key-based transformation technique for model protection.

Therefore, for the first time, in this paper, we propose a model protection method with a secret key set in such a way that a stolen model cannot be used without a key set. Specifically, the proposed method preprocesses input images with a secret key set and trains a model by using such preprocessed images. The preprocessing technique used in the proposed method is a low-cost block-wise operation. In addition, the proposed method does not modify the network, and therefore, there is no overhead for both training and inference time. In an experiment, the performance of a model protected by the proposed method is demonstrated not only to be close to that of a non-protected one when the key set is correct but also to significantly drop upon using an incorrect key set. We make the following contributions in this paper.
\begin{itemize}
\item We demonstrate block-wise image transformation with a secret key to be effective for model protection.
\item We conduct extensive experiments on different datasets including ImageNet and carry out key estimation attacks and fine-tuning attacks.
\end{itemize}

The rest of this paper is structured as follows. Section~\ref{sec:related-work} presents related work on conventional model watermarking, its problems and learnable image encryption. Section~\ref{sec:proposed} puts forward the proposed model protection method. Experiments and results are presented in Section~\ref{sec:results}.
Section~\ref{sec:discussion} includes discussion and analysis, and Section~\ref{sec:conclusion} concludes this paper. This paper is an extension of the work in~\cite{pyone2020training}. 

\section{Related Work}
\label{sec:related-work}
We review the existing model watermarking schemes of image classifiers and discuss problems with them. In addition, we also overview learnable encryption methods by which the proposed method has been inspired.

\subsection{Model Watermarking}
Digital watermarking technology is widely used to combat copyright infringement for multimedia data~\cite{1998-IEEE-Swanson}. An owner embeds a watermark into multimedia content (such as images, audio, etc.). When the protected content is stolen, the embedded watermark is extracted and used to verify ownership. In a similar fashion, to prevent the illegal distribution of DNN models, digital watermarking techniques are used to embed watermarks into proprietary DNN models. There are mainly two scenarios in DNN model watermarking: white-box and black-box.

A model watermarking scenario in white-box settings requires access to model weights for embedding and extracting a watermark. Uchida et al.\ first proposed a white-box model-watermarking method~\cite{2017-ICMR-Uchida}. A watermark is embedded in one or more layers of model weights by using ``an embedding regularizer,'' which is an additional regularization term in the loss function during training. Similarly, there are other works that follow the use of an additional regularization term as in~\cite{2018-Arxiv-Chen, 2018-Arxiv-Rouhani, 2019-NIPS-Fan}.

Extracting watermarks in white-box settings requires access to the model weights. To overcome this limitation, another model watermarking scenario for black-box settings was proposed, where an inspector observes the input and output of a model in doubt to verify the ownership of the model. In the black-box scenario, adversarial examples are exploited as a backdoor trigger set~\cite{2018-USENIX-Yossi, 2018-ACCCS-Zhang}, or a set of training examples is utilized so that a watermark pattern can be extracted from the inference of a model by using a specific set of training examples~\cite{2019-NIPS-Fan, 2019-MIPR-Sakazawa, 2020-NCA-Le}. Therefore, access to the model weights is not required to verify ownership in black-box settings.

The above-mentioned model-watermarking schemes focus on ownership verification. Thus, a stolen model can be directly used and exploited without arousing suspicion because the performance of a protected model (i.e., fidelity) is independent of the embedded watermark. \RC{In contrast, the proposed model protection is not a watermarking method, and it is more relevant to authorization or digital rights management because only the rightful user who has the correct key set can use a model to full capacity. Although frameworks for model watermarking and the proposed model protection are different, they are both necessary for dealing with digital rights management in different applications.}

\subsection{Model Watermarking with Passports}
Fan et.\ al~\cite{2019-NIPS-Fan} pointed out that conventional ownership verification schemes are vulnerable against ambiguity attacks~\cite{1998-IEEEJSAC-Craver} where two watermarks can be extracted from the same protected model, causing confusion regarding ownership. Therefore, Fan et.\ al~\cite{2019-NIPS-Fan} introduced passports and passport layers, which allow us to verify ownership with the correct passports. However, the passport in~\cite{2019-NIPS-Fan} is a set of extracted features of a secret image/images or equivalent random patterns from a pre-trained model. In addition, a network has to be modified with additional passport layers to use passports. Therefore, there are significant overhead costs in both the training and inference phases, in addition to user-unfriendly management of lengthy passports in~\cite{2019-NIPS-Fan}.

In this paper, we aim to protect a model by embedding a secret key with minimal impact on model performance. Similar to the work in~\cite{2019-NIPS-Fan}, a correct key is required for correct inference. However, the proposed method does not introduce any overhead in training or inference. Ownership is automatically verified upon being given the correct key.

\subsection{Learnable Image Encryption}
Learnable image encryption (LIE) is to perceptually encrypt images to mainly protect visual information on plain images while maintaining the network ability to learn the encrypted ones for classification tasks. \RC{Conventional LIE methods are classified into two classes in terms of application: LIE for privacy-preserving deep learning~\cite{2018-ICCETW-Tanaka,madono2020block,2019-Access-Warit,2019-ICIP-Warit,sirichotedumrong2020gan,ito2020image,ito2020framework} and LIE for adversarial robustness~\cite{2020-ICIP-Maung,2020-Arxiv-Maung}.}

\RC{LIE methods for privacy-preserving have two requirements: protecting visual information and maintaining a high classification accuracy under the use of encrypted images.}
In a block-wise manner, a color image is divided into blocks, and each block is processed by using a series of encryption with a common key to all blocks~\cite{2018-ICCETW-Tanaka} or with different keys~\cite{madono2020block}. In a pixel-wise manner, negative/positive transformation to each pixel and color shuffling across three channels are exploited to produce learnable encrypted images~\cite{2019-Access-Warit,  2019-ICIP-Warit}.
In contrast, a transformation network is trained in cooperation with a pre-trained classification model to generate images without visual information on plain images. One such work utilized a generative adversarial network (GAN)~\cite{sirichotedumrong2020gan}. To improve classification accuracy and robustness against various attacks, transformation networks have been proposed that use U-Net as in~\cite{ito2020image,ito2020framework}.

\RC{LIE methods for adversarial defenses~\cite{2020-ICIP-Maung,2020-Arxiv-Maung} have three requirements: a high classification accuracy, robustness against adversarial attacks, and resistance to key estimation attacks. The methods in this class do not aim to protect the visual information of plain images. Instead, a key is used to control the model's decision. In this paper, we do not propose a new encryption method. We adopt the methods in~\cite{2020-ICIP-Maung,2020-Arxiv-Maung} for a new application, model protection, for the first time. The proposed model protection method is carried out on the basis of LIE methods for adversarial defenses~\cite{2020-ICIP-Maung,2020-Arxiv-Maung}, but hyperparameters are carefully tuned for model protection purposes. LIE methods have never been applied to model protection applications. The contribution in this paper is to introduce some conventional image encryption algorithms into a model protection task.}

\section{Proposed Model-protection Method}
\label{sec:proposed}
\subsection{Notation}
The following notations are utilized throughout this paper.
\begin{itemize}
\item $w$, $h$, and $c$ are used to denote the width, height, and number of channels of an image.
\item The tensor $x \in {[0, 1]}^{c \times h \times w}$ represents an input color image.
\item The tensor $x' \in {[0, 1]}^{c \times h \times w}$ represents a transformed image.
\item $M$ is the block size of an image.
\item Tensors $x_b, x_b' \in {[0, 1]}^{h_b \times w_b \times p_b}$ are a block image and a transformed block image, respectively, where $w_b = \frac{w}{M}$ is the number of blocks across width $w$, $h_b = \frac{h}{M}$ is the number of blocks across height $h$, and $p_b = M \times M \times c$ is the number of pixels in a block.
\item A pixel value in a block image ($x_b$ or $x_b'$) is denoted by $x_b(i, j, k)$ or $x_b'(i, j, k)$, where $i \in \{0, \dots, h_b - 1\}$, $j \in \{0, \dots, w_b - 1\}$, and $k \in \{0, \dots, p_b - 1\}$ are indices corresponding to the dimension of $x_b$ or $x_b'$.
\item $B$ is a block of an image, and its dimension is $M \times M \times c$.
\item $\hat{B}$ is a flattened version of block $B$, and its dimension is $1 \times 1 \times p_b$.
\item $K$ denotes a set of keys
\item A password required for format-preserving encryption, which refers to encrypting in such a way that the output is in the same format as its input, is denoted as $password$.
\item $\text{Enc}(n, password)$ denotes format-preserving Feistel-based encryption~\cite{2010-NIST-Bellare} with a length of $3$, where $n$ is an integer (used only in FFX encryption).
\item An image classifier is denoted as $f(\cdot)$.
\end{itemize}

\subsection{Requirements of Proposed Scheme}
We consider a model protection scenario that aims to fulfill the following requirements:
\begin{enumerate}
\item Usability: A rightful user with key set $K$ can access a model without any noticeable overhead in both training and inference time, and performance degradation. The key management should be easy.
\item Unusability: Ideally, stolen models should not be usable in any case without key set $K$. In addition, even when the adversary retrains a stolen model with a forged key set, the performance of the model should heavily drop.
\end{enumerate}

\subsection{Overview}
\begin{figure}[!ht]
\centering\includegraphics{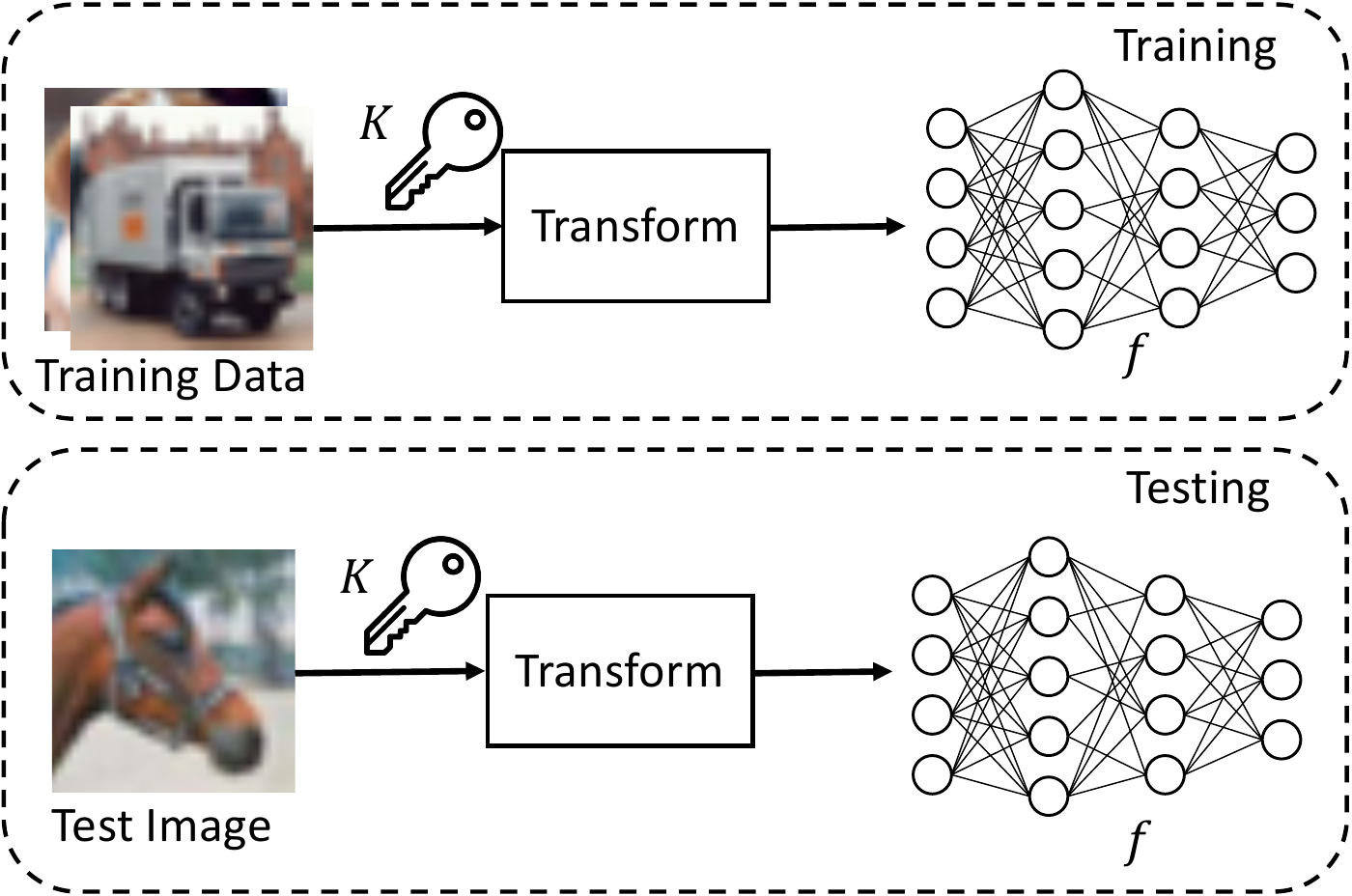}
\caption{\RA{Overview of image classification with proposed model protection method.}\label{fig:overview}}
\end{figure}

An overview of image classification with the proposed method is depicted in Fig.~\ref{fig:overview}. In the proposed model protection, input images are transformed by using secret key set $K$ before training or testing a model. Model $f$ is trained by using the transformed images. To test a trained model, test images are also transformed with the same key set $K$ before testing.

\begin{figure*}[!t]
\centering\includegraphics[width=\linewidth]{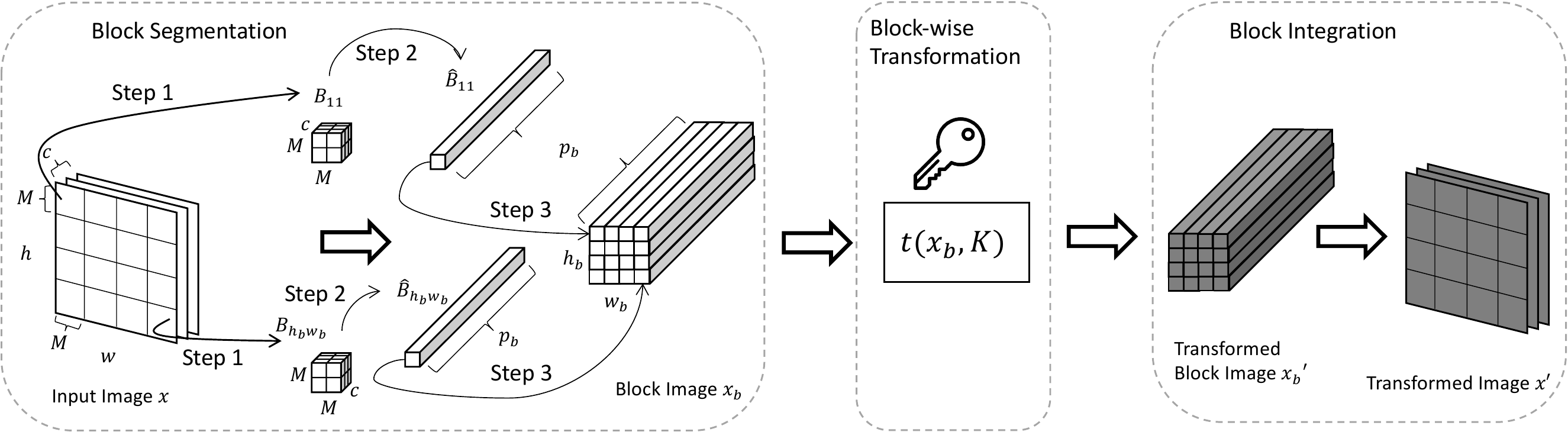}
\caption{\RA{Process of block-wise transformation.}\label{fig:general}}
\end{figure*}

The block-wise transformation consists of three steps: block segmentation, block-wise transformation, and block integration (see Fig.~\ref{fig:general}). The process of the block-wise transformation is shown as follows.

\begin{enumerate}
\item \textbf{Block Segmentation:} The process of block segmentation is illustrated in Fig.~\ref{fig:general}.
\begin{itemize}
\item Step 1: An input image $x$ is divided into blocks such that $\{B_{11}, B_{12}, \dots, B_{h_{b}w_{b}}\}$.
\item Step 2: Each block in $x$ is flattened to obtain \\$\{\hat{B}_{11}, \hat{B}_{12}, \dots, \hat{B}_{h_{b}w_{b}}\}$.
\item Step 3: The flattened blocks are concatenated in such a way that the relative spatial location among blocks in $x_b$ is the same as that among blocks in $x$.
\end{itemize}
\item \textbf{Block-wise Transformation:} Given a key set $K$, $x_b$ is transformed by using a block-wise transformation algorithm, $t(x_b, K)$. The transformed block image is written as
\begin{equation}
 x_b' = t(x_b, K). \label{eq:proposed}
\end{equation}
\item \textbf{Block Integration:} The transformed blocks in $x_b'$ are integrated back to the original dimension (i.e., $c \times h \times w$) in the reverse order to the block segmentation process for obtaining a transformed image $x'$.
\end{enumerate}

\subsection{Block-wise Transformation with Secret Key}
We introduce three block-wise transformations: pixel shuffling (SHF), negative/positive transformation (NP), and format-preserving Feistel-based encryption (FFX) for realizing $t(x_b, K)$. \RA{We selected these three transformations from among LIE methods for model protection applications because the requirements of LIE methods for these applications are expected to be similar to those for adversarial defenses.}

A set of keys $K$ consists of one or more keys depending on the desired number of transformations. For example, if one transformation (SHF) is used, $K = \{\alpha\}$, if two (SHF and NP) are used, $K = \{\alpha, \beta\}$, and if three (SHF, NP, and FFX) are used, $K = \{\alpha, \beta, \gamma\}$, where $\alpha$ is for SHF, $\beta$ is for NP, and $\gamma$ is for FFX, respectively.\@

Key $\alpha$ is a permutation vector, and it is defined as
  \begin{equation}
    \alpha = [\alpha_1,.,\alpha_k,.,\alpha_{k'}, \dots, \alpha_{p_b}], \alpha_k \in \{1,\ldots, p_b\},
  \end{equation}
where $\alpha_k \neq \alpha_{k'}$ if $k \neq k'$.

Key $\beta$ is a binary vector, and it is given by
  \begin{equation}
  \beta = [\beta_1, \dots, \beta_k, \dots, \beta_{p_b}], \beta_k \in \{0, 1\},
  \end{equation}
 where the value of the occurrence probability $P(\beta_k)$ is $0.5$.

Key $\gamma$ is defined as
  \begin{equation}
  \gamma = [\gamma_1, \dots, \gamma_k, \dots, \gamma_{p_b}], \gamma_k \in \{0, 1\},
  \end{equation}
where the value of the occurrence probability $P(\gamma_k)$ is $0.5$.

When SHF is used, $x_b'$ is obtained as:
 \begin{equation}
 x_b'(i, j, \alpha_k) = x_b(i, j, k).
 \end{equation}

When NP is used, every pixel value in $x_b$ needs to be at $255$ scale with 8 bits (i.e., multiply $x_b$ by $255$), and $x_b'$ is obtained as:
 \begin{equation}
 x_b'(i, j, k) = \left\{
 \begin{array}{ll}
 x_b(i, j, k) & (\beta_k = 0)\\
 x_b(i, j, k) \oplus (2^L - 1) & (\beta_k = 1),
 \end{array}
 \right.
 \end{equation}
where $\oplus$ is an exclusive or (XOR) operation, $L$ is the number of bits used in $x_b(i, j, k)$, and $L = 8$ is used in this paper. After the transformation, every pixel value in $x_b'$ is converted back to $[0, 1]$ scale (i.e., divide $x'_b$ by $255$).

When FFX is used, every pixel value in $x_b$ also needs to be at $255$ scale with 8 bits (i.e., multiply $x_b$ by $255$). In addition, FFX also requires a $password$ for format-preserving Feistel-based encryption~\cite{2010-NIST-Bellare}, and $x_b'$ is obtained as:
 \begin{equation}
 x_b'(i, j, k) = \left\{
 \begin{array}{ll}
 x_b(i, j, k) & (\gamma_k = 0)\\
 \text{Enc}(x_b(i, j, k), password) & (\gamma_k = 1).
 \end{array}
 \right.
 \end{equation}
Note that FFX~\cite{2010-NIST-Bellare} takes an integer value and outputs an integer; therefore, pixel values should be at $[0,255]$ scale. The pixel value $x_b(i,j,k) \in \{0, 1, \dots, 254, 255\}$ is encrypted by FFX with a length of $3$ digits to cover the whole range from $0$ to $255$. Therefore, FFX transforms each pixel with an integer value of ($[0,255]$ scale) into a pixel with an integer value of ($[0,999]$ scale), preserving the integer format. Another notable thing is that a $password$ for FFX can be arbitrary, and the only important thing is the location of the encrypted pixels that are determined by the key $\gamma$.

\begin{figure*}[!t]
\centering
\subfloat[Original]{\includegraphics[width=0.14\linewidth]{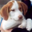}%
\label{fig:dog}}
\hfil
\subfloat[SHF]{\includegraphics[width=0.14\linewidth]{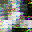}%
\label{fig:shf}}
\hfil
\subfloat[NP]{\includegraphics[width=0.14\linewidth]{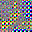}%
\label{fig:np}}
\hfil
\subfloat[FFX]{\includegraphics[width=0.14\linewidth]{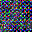}%
\label{fig:ffx}}
\hfil
\subfloat[SHF + NP]{\includegraphics[width=0.14\linewidth]{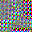}%
\label{fig:shfnp}}
\hfil
\subfloat[SHF + FFX]{\includegraphics[width=0.14\linewidth]{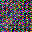}%
\label{fig:shfffx}}
\hfil
\subfloat[SHF + NP + FFX]{\includegraphics[width=0.14\linewidth]{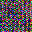}%
\label{fig:shfnpffx}}
\caption{\RA{Example of block-wise transformed images ($M = 4$) with key set $K$.}\label{fig:images}}
\end{figure*}

The overall block-wise transformation is detailed in Algorithm~\ref{algo:transform}. An example of images transformed by different transformations is shown in Fig.~\ref{fig:images}. \RA{The three transformations (SHF, NP, and FFX) were confirmed to have different performances in terms of classification accuracy and key estimation attack in~\cite{2020-ICIP-Maung,2020-Arxiv-Maung}, so these transformations are compared again under model protection in this paper. In particular, parameter $M$ affects the classification accuracy and resistance to key estimation attacks.}

\begin{algorithm}
\caption{Block-wise Transformation with Secret Key\label{algo:transform}}
\begin{algorithmic}[1]
\renewcommand{\algorithmicrequire}{\textbf{Input:}}
\renewcommand{\algorithmicensure}{\textbf{Output:}}
\REQUIRE{$x, K$}
\ENSURE{$x'$}
\STATE{Divide $x$ into blocks, $\{B_{11}, \ldots, B_{h_b w_b}\}$}
\STATE{Flatten blocks to vectors, $\{\hat{B}_{11}, \ldots, \hat{B}_{h_b w_b}\}$}
\STATE{Concatenate flattened blocks to obtain $x_b$}
\STATE{// \textit{To transform $x_b$ given $K$}}
\IF{SHF}
\STATE{$x_b' \leftarrow x_b[:, :, \alpha]$}
\ENDIF

\IF{NP}
\STATE{// \textit{Make pixel values be at 255 scale}}
\STATE{$x_b \leftarrow x_b \cdot 255$}
\STATE{$x_b'[:, :, \beta] \leftarrow 255 - x_b[:, :, \beta]$}
\STATE{$x_b' \leftarrow x_b' / 255$}
\ENDIF

\IF{FFX}
\STATE{// \textit{Make pixel values be at 255 scale}}
\STATE{$x_b \leftarrow x_b \cdot 255$}
\STATE{$x_b'[:, :, \gamma] \leftarrow \text{Enc}(x_b[:, :, \gamma], password$)}
\STATE{max $\leftarrow$ the maximum value of the encryption}
\STATE{$x_b' \leftarrow x_b' / \text{max}$}
\ENDIF

\STATE{$x' \leftarrow$ Integrate blocks in $x_b'$}
\end{algorithmic}
\end{algorithm}

\subsection{Robustness Against Attacks}
\label{sec:attacks}
A threat model includes a set of assumptions such as an attacker's goals, knowledge, and capabilities. An attacker may steal a model to achieve different goals. In this paper, we consider the attacker's goal to be to make use of a stolen model by estimating a key set or fine-tuning the stolen model for different purposes. In this regard, we assume that the attacker knows the transformation details such as the block size and type of transformation, and a small subset of the training dataset. Therefore, the attacker may observe the accuracy of his or her test dataset to estimate a key set or fine-tune the stolen model.
We carry out the following possible attacks with the intent of stealing a model to evaluate the robustness of the proposed method. In experiments, the proposed method will be demonstrated to be robust against attacks.

\subsubsection{Key Estimation Attack}
\label{sec:estimation}
We consider a scenario where a model is stolen and transformation details are known except the secret key. The key may be estimated by brute-force checking possible keys. The key space $\mathcal{K}$ of each transformation is given by
\begin{equation}
  \mathcal{K}_{\alpha}(c \times M \times M) = (c \times M \times M)! \;\;\;\text{(SHF)}, \label{eqn:shf}
\end{equation}
\begin{equation}
  \mathcal{K}_{\beta}(c \times M \times M) = 2^{(c \times M \times M)} \;\;\;\text{(NP), and} \label{eqn:np}
\end{equation}
\begin{equation}
  \mathcal{K}_{\gamma}(c \times M \times M) = 2^{(c \times M \times M)} \;\;\;\text{(FFX).} \label{eqn:ffx}
\end{equation}
Therefore, the key space will vary with respect to block size $M$ and the type of block-wise transformation used for protecting a model.

The attacker may estimate the key heuristically by observing the accuracy over a batch of images. \RA{Algorithm~\ref{algo:key-estimation} describes the process of estimating key set $K$. First, we randomly initialize key set $K' = \{\alpha', \beta', \gamma'\}$ in accordance with the transformation. Next, we also initialize a set of index pairs $\mathcal{P}$ as $\mathcal{P} = \{(1, 2), (1, 3), \ldots, (c \times M \times M - 1, c \times M \times M)\}$ for keys $\alpha'$, $\beta'$, or $\gamma'$. The number of all possible combinations of pairs for each key can be computed as a binomial coefficient given by}
\begin{equation}
  \RA{\left| \mathcal{P} \right| = {}_n C_r = \frac{n!}{r!(n-r)!},}
\end{equation}
\RA{where $n = c \times M \times M$, and $r = 2$. For each index pair, we swap the pair in $\alpha'$, $\beta'$, or $\gamma'$ if the swap improves the accuracy as shown in Algorithm~\ref{algo:key-estimation}.}

Key estimation attacks do not guarantee that the attacker will find the correct key because the attacker does not know the actual performance of the correct key. However, the attacker may perform fine-tuning attacks to exploit a stolen model as below.

\begin{algorithm}
  \caption{\RA{Key Estimation}\label{algo:key-estimation}}
\begin{algorithmic}[1]
\renewcommand{\algorithmicrequire}{\textbf{Input:}}
\renewcommand{\algorithmicensure}{\textbf{Output:}}
\REQUIRE{Input images with labels}
\ENSURE{$K'$}
\STATE{Initialize $K' = \{\alpha', \beta', \gamma'\}$ in accordance with the transformation}
\STATE{Initialize $\mathcal{P} = \{(1, 2), (1, 3), \ldots, (i, j), \ldots, (c \times M \times M - 1, c \times M \times M)\}$}
\STATE{accuracy $\leftarrow$ Calculate accuracy of input images}
\FOR{Each index pair $(i, j)$ in $\mathcal{P}$}
\IF{SHF}
\STATE{$(\alpha'_{i}, \alpha'_{j}) \leftarrow (\alpha'_{j}, \alpha'_{i})$}
\STATE{current\_accuracy $\leftarrow$ Calculate accuracy of input images}
\IF{current\_accuracy $>$ accuracy}
\STATE{accuracy $\leftarrow$ current\_accuracy}
\ELSE
\STATE{$(\alpha'_{i}, \alpha'_{j}) \leftarrow (\alpha'_{j}, \alpha'_{i})$}
\ENDIF
\ENDIF
\IF{NP}
\STATE{$(\beta'_{i}, \beta'_{j}) \leftarrow (\beta'_{j}, \beta'_{i})$}
\STATE{current\_accuracy $\leftarrow$ Calculate accuracy of input images}
\IF{current\_accuracy $>$ accuracy}
\STATE{accuracy $\leftarrow$ current\_accuracy}
\ELSE
\STATE{$(\beta'_{i}, \beta'_{j}) \leftarrow (\beta'_{j}, \beta'_{i})$}
\ENDIF
\ENDIF
\IF{FFX}
\STATE{Swap $\gamma'_{i}$ with $\gamma'_{j}$ in key $\gamma'$}
\STATE{$(\gamma'_{i}, \gamma'_{j}) \leftarrow (\gamma'_{j}, \gamma'_{i})$}
\STATE{current\_accuracy $\leftarrow$ Calculate accuracy of input images}
\IF{current\_accuracy $>$ accuracy}
\STATE{accuracy $\leftarrow$ current\_accuracy}
\ELSE
\STATE{$(\gamma'_{i}, \gamma'_{j}) \leftarrow (\gamma'_{j}, \gamma'_{i})$}
\ENDIF
\ENDIF
\ENDFOR
\end{algorithmic}
\end{algorithm}

\subsubsection{Fine-tuning Attack with Incorrect Key Set and Small Dataset}
Fine-tuning (transfer learning)~\cite{2015-ICLR-Simonyan} is to train a model on top of pre-trained weights. Since fine-tuning alters the weights of the model, an attacker may use fine-tuning as an attack to overwrite a protected model with the intent of forging keys. \RA{The goal of this attack is to replace the key set with a different key set by retraining a protected model with a small subset of a dataset.}
We can consider such an attack scenario where the adversary has a subset of dataset $D'$ and retrains the model with a forged key set ($K'$).

\subsubsection{Fine-tuning Attack with New Dataset}
\RA{We assume an attacker may steal a protected model and fine-tune the model with a new dataset. The goal of this attack is to replace a protected model with an unprotected one without any key by using transfer learning.}

In practice, CNNs are not trained from the beginning with random weights because creating a large dataset like ImageNet is difficult and expensive. Therefore, CNNs are usually pre-trained with a larger dataset (e.g., ImageNet), known as transfer learning~\cite{2015-ICLR-Simonyan}. There are two major transfer-learning scenarios:
\begin{itemize}
\item \textbf{Fixed CNN:} A pre-trained CNN model is used as a fixed feature extractor, and the last fully connected layer is replaced with a targeted number of classes. In other words, convolutional layers are frozen, and only the last fully connected layer is trained with random initialization from scratch.
\item \textbf{Fine-tuned CNN:} In this scenario, the CNN is fine-tuned from a pre-trained model. Here, it is possible that some convolutional layers can be fixed or the whole CNN is fine-tuned.
\end{itemize}

\section{Experiments and Results}
\label{sec:results}
To verify the effectiveness of the proposed model protection method, we ran a number of experiments on different datasets. All the experiments were carried out in PyTorch~\cite{pytorch} platform.

\subsection{Datasets}
We conducted image classification experiments on datasets with different scales, namely, the CIFAR (both 10 and 100 classes)~\cite{2009-Report-Krizhevsky} and ImageNet~\cite{ILSVRC15} datasets.

For the CIFAR-10 and CIFAR-100 datasets, we used a batch size of 128 and live augmentation (random cropping with padding of 4 and random horizontal flip) on training sets. Both datasets consist of 60,000 color images (dimension of $32 \times 32 \times 3$) where 50,000 images are for training and 10,000 for testing. There are 10 classes (6000 images for each class) for the CIFAR-10 dataset and 100 classes (600 images for each class) for the CIFAR-100 dataset.

ImageNet comprises 1.28 million color images for training and 50,000 color images for validation. We progressively resized images during training, starting with larger batches of smaller images to smaller batches of larger images. We adapted three phases of training from the DAWNBench top submissions as mentioned in~\cite{2020-ICLR-Wong}. Phases 1 and 2 resized images to 160 and 352 pixels, respectively, and phase 3 used the entire image size from the training set. The augmentation methods used in the experiment were random resizing, cropping (sizes of $128$, $224$, and $288$, respectively, for each phase), and random horizontal flip.

\subsection{Networks}
We utilized deep residual networks~\cite{2016-CVPR-He} with 18 layers (ResNet18) and trained for $200$ epochs with cyclic learning rates~\cite{2017-Arxiv-Smith} and mixed precision training~\cite{2017-Arxiv-Micikevicius} for the CIFAR datasets. The parameters of the stochastic gradient descent (SGD) optimizer were a momentum of $0.9$, weight decay of $0.0005$, and maximum learning rate of $0.2$. For the ImageNet dataset, we used ResNet50 with pre-trained weights. We adapted the training settings from~\cite{2020-ICLR-Wong} with the removal of weight decay regularization from batch normalization layers. The network was trained for 15 epochs in total for the ImageNet dataset.

\subsection{Classification Performance}
\RA{We trained protected models by using images transformed by various transformations (both single and combined transformations) with different block sizes (i.e., $M \in \{2, 4, 8, 16\}$) on three different datasets (CIFAR-10, CIFAR-100, and ImageNet).}
The models are named after the shorthand of the respective transformations. For example, the model trained by using images transformed by SHF transformation is denoted as SHF, that by NP transformation as NP, and so on.
We tested the protected models under three conditions: with correct key set $K$, with incorrect key set $K'$, and with plain images (without any transformation).

\RC{In the experiments, the correct keys were generated by using a random number generator from the PyTorch platform~\cite{pytorch} with a seed value of 42 (64-bit integer), and we used a publicly available library for FFX (pyffx~\cite{pyffx}) with the password string ``password.'' The incorrect keys were also generated by using the same random number generator from the PyTorch platform~\cite{pytorch} with 1000 random seed values (64-bit integer).}

Table~\ref{tab:results} summarizes the simulation results for all three datasets, \RC{where the classification accuracy for incorrect key set $K'$ was averaged over 1000 random key sets.}

\RA{\textbf{SHF:} For SHF, as block size $M$ was increased, the classification accuracy decreased when using correct key set $K$.
Under the use of incorrect key set $K'$ or plain images, the accuracy significantly dropped as block size $M$ was increased, suggesting resistance against unauthorized access.
Therefore, the selection of $M$ controls the trade-off between classification accuracy and resistance against illegal usage.}

\RA{\textbf{NP, FFX:} For NP, the highest accuracy values were achieved for all transformations on all the datasets for each block size with correct key set $K$.
When using $M = 8$ or $16$ on the CIFAR-10 and CIFAR-100 datasets, NP had the highest accuracy, and FFX had the second highest accuracy under the use of incorrect key set $K'$ or plain images except for $M = 8$ on the CIFAR-100 dataset.
In contrast, on ImageNet, the accuracy was significantly low for both NP and FFX under the use of incorrect key set $K'$ or plain images because the size of the images in the ImageNet dataset is larger than that of CIFAR-10 and CIFAR-100.}

\RA{\textbf{Combined:} The models with combined transformations such as SHF + NP and SHF + FFX decreased the classification accuracy, compared with those with NP and FFX, when using $M = 8$ or $16$. Under the use of incorrect key set $K'$ or plain images, the accuracy severely dropped for all block sizes. The advantage of a combined transformation is that it can increase the key space, but it slightly reduces the classification accuracy.}

\RA{From the empirical results, generally, the performance of an incorrect key set depends on the number of classes in a dataset. When using a dataset with a large number of classes such as CIFAR-100 and ImageNet, the accuracy of the incorrect key set was low due to difficulty in classifying images transformed by using an incorrect key set or plain images.} In summary, the proposed method had a high classification accuracy (i.e., close to baseline accuracy) when correct key set $K$ was given. In contrast, the accuracy deteriorated significantly when using incorrect key set $K'$ or plain images. \RA{Since models with $M = 4$ provided a good trade-off between classification accuracy and resistance against unauthorized access, we focused on $M = 4$ in the following sections to further evaluate against attacks on the CIFAR-10 and CIFAR-100 datasets.}

\robustify\bfseries
\sisetup{table-parse-only,detect-weight=true,detect-inline-weight=text,round-mode=places,round-precision=2}
\begin{table*}[htbp]
\centering
\caption{\RA{Accuracy (\SI{}{\percent}) of protected models and baseline model for three datasets. Best results are in \textbf{bold}.}\label{tab:results}}
\begin{tabular}{cl|SSS|SSS|SSS}

\toprule
                               && \multicolumn{3}{c|}{CIFAR-10} & \multicolumn{3}{c|}{CIFAR-100} & \multicolumn{3}{c}{ImageNet}\\
                               &{Model} & {Correct} & {Incorrect} & {Plain} & {Correct} & {Incorrect} & {Plain} & {Correct} & {Incorrect} & {Plain}\\
                               && {($K$)} & {($K'$)} & & {($K$)} & {($K'$)} & & {($K$)} & {($K'$)} & \\
\midrule
 \multirow{6}{*}{\rot{$M = 2$}}&{SHF} & 94.76  & 36.36 & 31.43 & 77.03 & 8.6 & 6.09 & 73.00 & 46.57 & 40.35\\
                               &{NP} & \bfseries \num{95.32} & 18.44 & 13.91 & \bfseries \num{77.88} & 2.19 & \bfseries \num{1.19} & \bfseries \num{73.04} & 6.53 & 0.98\\
                               &{FFX} & 93.80 & 15.69 & 38.84 & 74.36 & 3.41 & 11.50 & 72.43 & \bfseries \num{0.12} & 0.23\\
                               &{SHF + NP} & 94.50 & 20.65 & \bfseries \num{11.79} & 76.76 & 2.52 & 1.49 & 72.90 & 4.87 & 1.12\\
                               &{SHF + FFX} & 93.02 & 15.3 & 19.60 & 73.17 & 2.77 & 5.24 & 72.30 & 0.47 & 0.19\\
                               &{SHF + NP + FFX} & 92.82 & \bfseries \num{14.03} & 18.69 & 73.08 & \bfseries \num{1.46} & 1.94 & 71.96 & 0.16 & \bfseries \num{0.18}\\
\midrule
 \multirow{6}{*}{\rot{$M = 4$}}&{SHF} & 92.58 & 20.23 & 27.77 & 72.05 & 4.9 & 5.85 & 72.41 & 13.06 & 32.98\\
                               &{NP} & \bfseries \num{93.41} & 12.67 & \bfseries \num{12.17} & \bfseries \num{73.11} & 1.32 & 1.62 & \bfseries \num{72.63} & 0.68 & 0.36\\
                               &{FFX} & 92.29 & 18.38 & 37.06 & 69.92 & 3.9 & 12.73 & 72.17 & 0.15 & \bfseries \num{0.15}\\
                               &{SHF + NP} & 92.37 & 12.11 & 12.35 & 71.27 & 1.35 & 1.93 & 72.15 & 0.21 & 0.25\\
                               &{SHF + FFX} & 90.71 & 12.31 & 20.75 & 68.48 & 1.85 & 4.16 & 71.96 & 0.14  & 0.17\\
                               &{SHF + NP + FFX} & 90.50 & \bfseries \num{10.6} & 13.10 & 68.20 & \bfseries \num{1.09} & \bfseries \num{1.53} & 71.68 & \bfseries \num{0.12} & 0.16\\
\midrule
 \multirow{6}{*}{\rot{$M = 8$}}&{SHF} & 86.40 & 17.0 & 14.42 & 62.18 & 2.2 & 2.87 & 70.85 & 1.25 & 11.74\\
                               &{NP} & 91.54 & 71.35 & 79.51 & 67.07 & 1.8 & 1.65 & \bfseries \num{71.83} & 0.26 & 0.12\\
                               &{FFX} & \bfseries \num{92.00} & 47.07 & 37.25 & \bfseries \num{69.60} & 9.66 & 11.12 & 71.46 & 0.3 & \bfseries \num{0.09}\\
                               &{SHF + NP} & 86.47 & 12.16 & 14.75 & 62.75 & 1.42 & 1.56 & 71.14 & 0.19 & 0.86\\
                               &{SHF + FFX} & 86.01 & 11.81 & 15.20 & 61.37 & 1.29 & 1.70 & 70.77 & 0.11 & 0.14\\
                               &{SHF + NP + FFX} & 85.49 & \bfseries \num{10.23} & \bfseries \num{10.31} & 60.96 & \bfseries \num{1.06} & \bfseries \num{1.02} & 70.18 & \bfseries \num{0.1} & 0.11\\
\midrule
 \multirow{6}{*}{\rot{$M = 16$}}&{SHF} & 77.24 & 10.57 & 13.36 & 50.87 & 1.39 & 1.36 & 67.03 & 0.23 & 4.22\\
                                &{NP} & \bfseries \num{92.68} & 88.27 & 89.00 & \bfseries \num{70.68} & 48.17 & 47.38 & \bfseries \num{70.19} & 0.97 & 5.52\\
                                &{FFX} & 91.38 & 72.91 & 29.35 & 69.78 & 34.92 & 9.19 & 69.24 & 2.07 & 0.14\\
                                &{SHF + NP} & 77.52 & 10.66 & 11.70 & 50.33 & \bfseries \num{1.0} & \bfseries \num{1.02} & 67.50 & 0.11 & 0.18\\
                               &{SHF + FFX} & 76.28 & 10.2 & 12.79 & 49.53 & 1.21 & 1.45 & 63.75 & 0.16 & 0.13\\
                               &{SHF + NP + FFX} & 75.78 & \bfseries \num{10.0} & \bfseries \num{9.92} & 49.63 & \bfseries \num{1.0} & 1.04 & 63.43 & \bfseries \num{0.09} & \bfseries \num{0.12}\\
\midrule
  & {Baseline} &\multicolumn{3}{c|}{95.45 (Not protected)} & \multicolumn{3}{c|}{77.67 (Not protected)} & \multicolumn{3}{c}{73.70 (Not protected)}\\
\bottomrule
\end{tabular}
\end{table*}

\subsection{Robustness Against Key Estimation Attack}
The proposed method was evaluated in a scenario involving a key estimation attack in accordance with Algorithm~\ref{algo:key-estimation}. \RA{As described in Section~\ref{sec:proposed}-\ref{sec:attacks}-\ref{sec:estimation}), elements in each key were rearranged in accordance with the improvement in accuracy, and the resulting estimated key set $K'$ was used to evaluate the performance of the protected models.}

\RA{Table~\ref{tab:key-estimation} captures the classification performance of the models under the use of estimated key set $K'$ on the CIFAR-10 and CIFAR-100 datasets. The estimated keys were not good enough to provide a reasonable accuracy except for FFX, which replaces a pixel value with a random value by using a password, so the encrypted pixel value contains almost no information. Therefore, the location of the un-encrypted pixel values plays an important role in the model’s decision-making process, as the encrypted pixel values are not important. This property helps an attacker to effectively find a good key when performing key estimation attacks (Algorithm~\ref{algo:key-estimation}). In contrast, in the other two transformations (SHF and NP), transformed pixel values have some information, and both positions of un-encrypted pixels and pixel values are important. Therefore, the indication to search for a good key was difficult for SHF and NP compared to FFX.\
}

\robustify\bfseries
\sisetup{table-parse-only,detect-weight=true,detect-inline-weight=text,round-mode=places,round-precision=2}
\begin{table}[htbp]
\centering
\caption{\RA{Accuracy (\SI{}{\percent}) of protected models ($M = 4$) under use of estimated key set $K'$}\label{tab:key-estimation}}
\begin{tabular}{lSS}

  \toprule
   & {CIFAR-10} & {CIFAR-100}\\
  {Model} & {Estimated ($K'$)} & {Estimated ($K'$)}\\
  \midrule
  {SHF} & 25.66 & 10.20\\
  {NP} & 37.44 & 8.78\\
  {FFX} & 80.97 & 49.18\\
  {SHF + NP} & 14.53 & 2.78\\
  {SHF + FFX} & 15.04 & 2.36\\
  {SHF + NP + FFX} & 11.00 & 1.47\\
  \bottomrule
\end{tabular}
\end{table}

\subsection{Robustness Against Fine-tuning Attack with Incorrect Key and Small Dataset}
We ran an experiment with different sizes for the adversary's dataset (i.e., $\left| D' \right| \in \{100, 500, 1000, 10000\}$) for the CIFAR-10 and CIFAR-100 datasets. We retrained the models with $D'$ for 30 epochs. Table~\ref{tab:fine-tune} shows the results of fine-tuning attacks for both datasets. \RA{Table~\ref{tab:results} shows the performance of the incorrect key set before the model weights were modified. In contrast, fine-tuning with $\left| D' \right| = 100$ modified the weights. Therefore, the models in Table~\ref{tab:results} are different from those in Table~\ref{tab:fine-tune}, so the accuracy was not equal to that of the incorrect key set in Table~\ref{tab:results}.}

Although the accuracy improved with respect to the size of the adversary's dataset, it was still lower than the performance of the correct key set $K$ as presented in Table~\ref{tab:results}. \RA{Therefore, the results show that the compromised models were not as good as the original models, even when fine-tuning with a small subset of a dataset. As a result, the attacker is not able to use the model to full capacity, suggesting robustness against this type of attack.}

\robustify\bfseries
\sisetup{table-parse-only,detect-weight=true,detect-inline-weight=text,round-mode=places,round-precision=2}
\begin{table*}[htbp]
\centering
\caption{\RA{Accuracy (\SI{}{\percent}) of protected models under fine-tuning attacks with incorrect key and small dataset}\label{tab:fine-tune}}
\resizebox{\linewidth}{!}{%
\begin{tabular}{l|S|S|S|S|S|S|S|S}

  \toprule

& \multicolumn{4}{c|}{CIFAR-10} & \multicolumn{4}{c}{CIFAR-100}\\
  {Model} & {$\left| D' \right| = 100$} & {$\left| D' \right| = 500$} & {$\left| D' \right| = 1000$} & {$\left| D' \right| = 10000$} & {$\left| D' \right| = 100$} & {$\left| D' \right| = 500$} & {$\left| D' \right| = 1000$} & {$\left| D' \right| = 10000$}\\
\midrule
  {SHF} & 12.69 & 38.33 & 46.73 & 86.31 & 2.65 & 8.52 & 12.30 & 58.99\\
  {NP} & 10.57 & 37.25 & 47.41 & 87.06 & 3.00 & 8.19 & 11.57 & 60.70\\
  {FFX} & 10.15 & 32.30 & 40.52 & 86.04 & 1.00 & 6.58 & 9.62 & 58.64\\
  {SHF + NP} & 14.03 & 37.50 & 46.36 & 85.37 & 2.70 & 7.88 & 11.77 & 57.80\\
  {SHF + FFX} & 12.54 & 46.28 & 55.11 & 83.74 & 2.54 & 18.21 & 29.83 & 56.14\\
  {SHF + NP + FFX} & 11.15 & 39.59 & 48.13 & 83.27 & 3.23 & 12.09 & 24.14 & 54.51\\
  \bottomrule
\end{tabular}
}
\end{table*}

\subsection{Robustness Against Fine-tuning Attack with New Dataset}
We simulated this attack scenario by fine-tuning the CIFAR-100 to the CIFAR-10 under both conditions (Fixed CNN and Fine-tuned CNN). We fine-tuned the CIFAR-100 model for 25 epochs. The parameters of the stochastic gradient descent (SGD) optimizer were a learning rate of 0.001 and a momentum of 0.9, and a StepLR scheduler was used with a step size of 7 and a gamma value of 0.1.

Table~\ref{tab:finetune-dataset} shows the results of fine-tuning the CIFAR-100 dataset to the CIFAR-10 dataset. \RA{Training only the last layer (Fixed CNN) did not provide good accuracy even for the non-protected models (Plain). However, for ``Fine-tuned CNN,'' the non-protected (Plain) model was fine-tuned to an accuracy of \SI{90.60}{\percent}, which was closer to the baseline accuracy (i.e., \SI{95.45}{\percent}). In contrast, the fine-tuned accuracies of the protected models were lower (\SI{83.46}{\percent} for SHF, \SI{83.52}{\percent} for NP, and \SI{86.18}{\percent} for FFX) than that of the plain model. For the fine-tuned CNN, the results show that the protected models were still transferable, although this was not as good as fine-tuning from the non-protected models.}

\robustify\bfseries
\sisetup{table-parse-only,detect-weight=true,detect-inline-weight=text,round-mode=places,round-precision=2}
\begin{table}[htbp]
\centering
\caption{\RA{Accuracy (\SI{}{\percent}) of protected models under fine-tuning attacks with new dataset (CIFAR-100 to CIFAR-10) for fixed CNN and fine-tuned CNN}\label{tab:finetune-dataset}}
\begin{tabular}{lSS}
  \toprule
{Model} & {Fixed} & {Fine-tuned}\\
{} & {CNN} & {CNN}\\
  \midrule
  {SHF} & 44.14 & 83.46\\
  {NP} & 36.70 & 83.52\\
  {FFX} & 52.29 & 86.18\\
  {SHF + NP} & 37.67 & 82.40\\
  {SHF + FFX} & 37.00 & 79.67\\
  {SHF + NP + FFX} & 26.47 & 71.10\\
  \midrule
  {Plain} & 72.85 & 90.60\\
  \bottomrule
\end{tabular}
\end{table}

\subsection{Comparison with State-of-the-art Methods}
\RA{Both watermarking and encryption algorithms can be used to protect the copyright of digital products, but the former is imperceptible, while the latter is a means of direct encryption. It is difficult to compare both approaches in terms of their aims and robustness against attacks. In particular, in conventional model watermarking methods, the embedded watermark is independent of model performance.} Therefore, we compared the proposed protected model (NP) with the state-of-the-art passport protected model, Scheme $\mathcal{V}_1$~\cite{2019-NIPS-Fan} in terms of classification accuracy with/without correct key/passports, overheads, network modification and key management, for both the CIFAR-10 and CIFAR-100 datasets. \RA{Scheme $\mathcal{V}_1$~\cite{2019-NIPS-Fan} was not trained and tested using the same settings as the proposed method because the network in $\mathcal{V}_1$ was modified with passport layers and the hyperparameters were based on the modified network. In contrast, the proposed method used a standard ResNet18 and was trained with cyclic learning rates~\cite{2017-Arxiv-Smith} and mixed precision training~\cite{2017-Arxiv-Micikevicius}.}

\textbf{CIFAR-10:} In terms of accuracy when the correct key/passport was given, the accuracy of $\mathcal{V}_1$ was slightly higher than that of NP at {\SI{1.21}{\percent}}. However, it was confirmed that if block size $M = 2$ was used, NP achieved higher accuracy than $\mathcal{V}_1$ (i.e., \SI{95.32}{\percent}). When estimated incorrect key set was given, the accuracy of NP significantly dropped. In contrast, when reverse-engineered (i.e., estimated) passports were used, the accuracy of $\mathcal{V}_1$ was high (\SI{70}{\percent}).

\textbf{CIFAR-100:} Similarly, for CIFAR-100, the accuracy of $\mathcal{V}_1$ was also slightly higher. However, when an estimated key was given, the proposed method was more resistant than $\mathcal{V}_1$.

In terms of overhead, $\mathcal{V}_1$ modifies a network with additional passport layers; therefore, it introduces a training and inference overhead for both datasets. \RA{The overheads in~\cite{2019-NIPS-Fan} are based on the relative recorded time taken as mentioned in the paper by the original authors.} In contrast, the proposed model NP does not have any noticeable overhead, and there is no modification in the network for both datasets~\cite{2019-NIPS-Fan}. \RA{Moreover, the block-wise transformation in the proposed model protection can be efficiently implemented with vectorized operations; therefore, pre-processing with the block-wise transformation does not cause any noticeable overheads in both training and testing.} From a key management perspective, $\mathcal{V}_1$ requires a trained model to generate passports, and the proposed model NP does not need any model to generate keys. Therefore, the key management of the proposed method is simple and straightforward.

\robustify\bfseries
\sisetup{table-parse-only,detect-weight=true,detect-inline-weight=text,round-mode=places,round-precision=2}
\begin{table*}[htbp]
\centering
\caption{Comparison of proposed protected model NP and state-of-the-art passport-protected model in terms of classification accuracy (\SI{}{\percent}) for CIFAR datasets\label{tab:comparison}}
\resizebox{\linewidth}{!}{%
\begin{tabular}{lSScccc}

\toprule
\multicolumn{7}{c}{CIFAR-10}\\
\midrule
{Model} & {Correct $K$ / Passports} & {Estimated $K'$ / Passports} & {Training} & {Inference} & {Network} & {Key}\\
& {} & {} & {Overhead} & {Overhead} & {Modification} & {Management}\\
\midrule
{NP (Proposed)} & 93.41 & 37.44 & {Negligible} & {Negligible} & {No} & {Easy}\\
\midrule
{Scheme $\mathcal{V}_1$}~\cite{2019-NIPS-Fan} & 94.62 & 70.00 & {15--30\SI{}{\percent}~\cite{2019-NIPS-Fan}} & {\SI{10}{\percent}~\cite{2019-NIPS-Fan}} & {Yes} & {Difficult}\\
\bottomrule
\multicolumn{7}{c}{CIFAR-100}\\
\midrule
{NP (Proposed)} & 73.10 & 8.78 & {Negligible} & {Negligible} & {No} & {Easy}\\
\midrule
{Scheme $\mathcal{V}_1$}~\cite{2019-NIPS-Fan} & 75.52 & 35.00 & {15--30\SI{}{\percent}~\cite{2019-NIPS-Fan}} & {\SI{10}{\percent}~\cite{2019-NIPS-Fan}} & {Yes} & {Difficult}\\
\bottomrule
\end{tabular}
}
\end{table*}

\begin{figure*}[!t]
\centering
\subfloat{\includegraphics[width=0.33\linewidth]{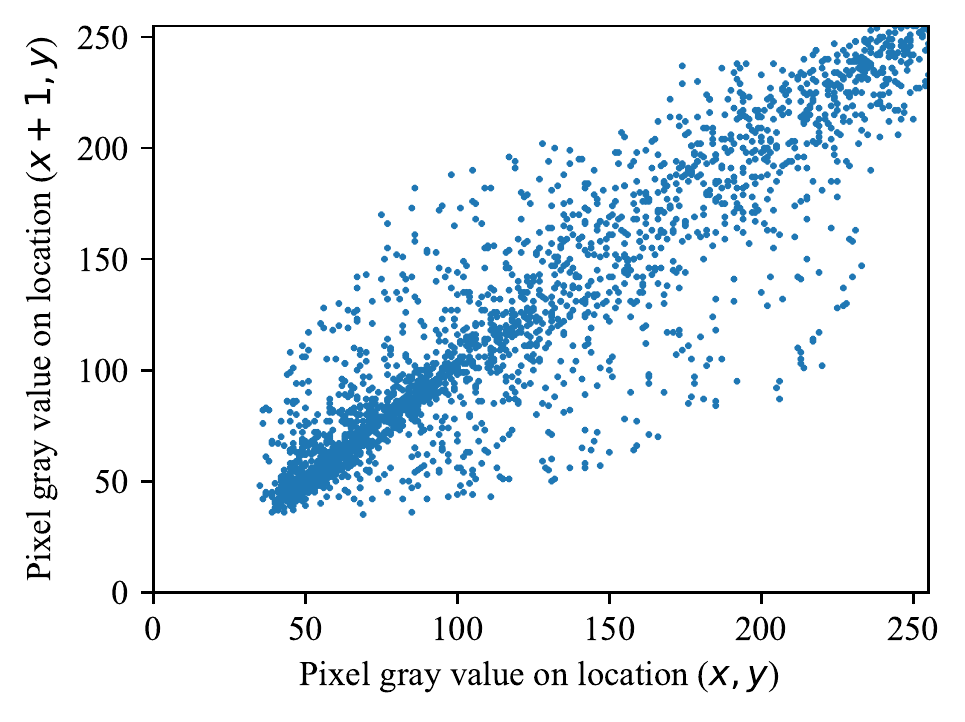}%
\label{fig:puppy_h}}
\hfil
\renewcommand{\thesubfigure}{a}
\subfloat[]{\includegraphics[width=0.33\linewidth]{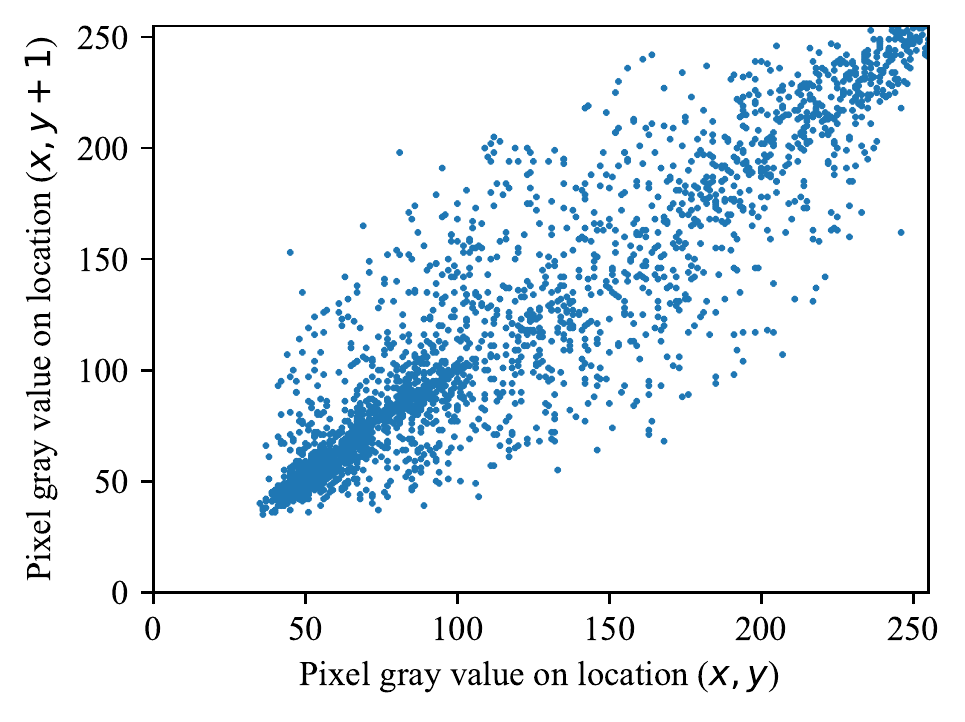}%
\label{fig:puppy_v}}
\hfil
\subfloat{\includegraphics[width=0.33\linewidth]{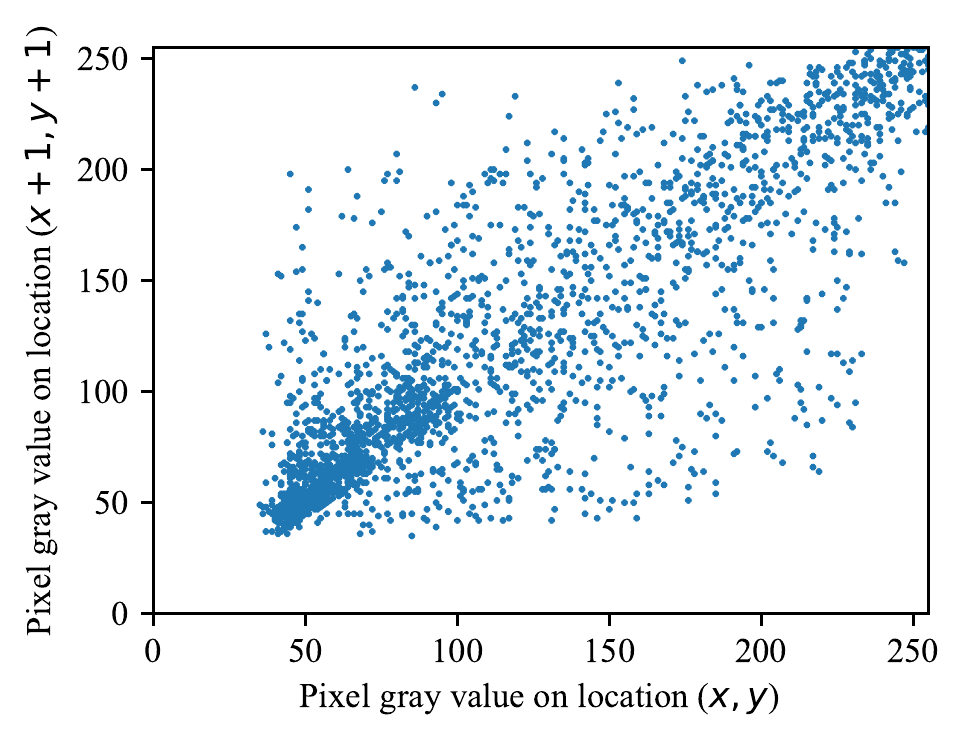}%
\label{fig:puppy_d}}\\
\subfloat{\includegraphics[width=0.33\linewidth]{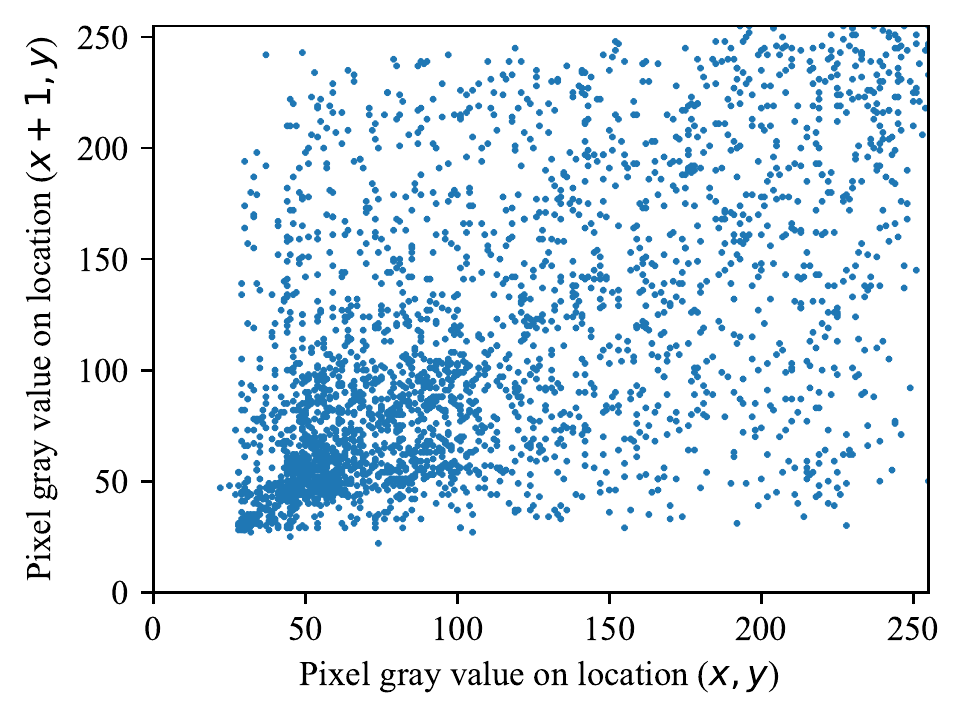}%
\label{fig:puppy_shf_h}}
\hfil
\renewcommand{\thesubfigure}{b}
\subfloat[]{\includegraphics[width=0.33\linewidth]{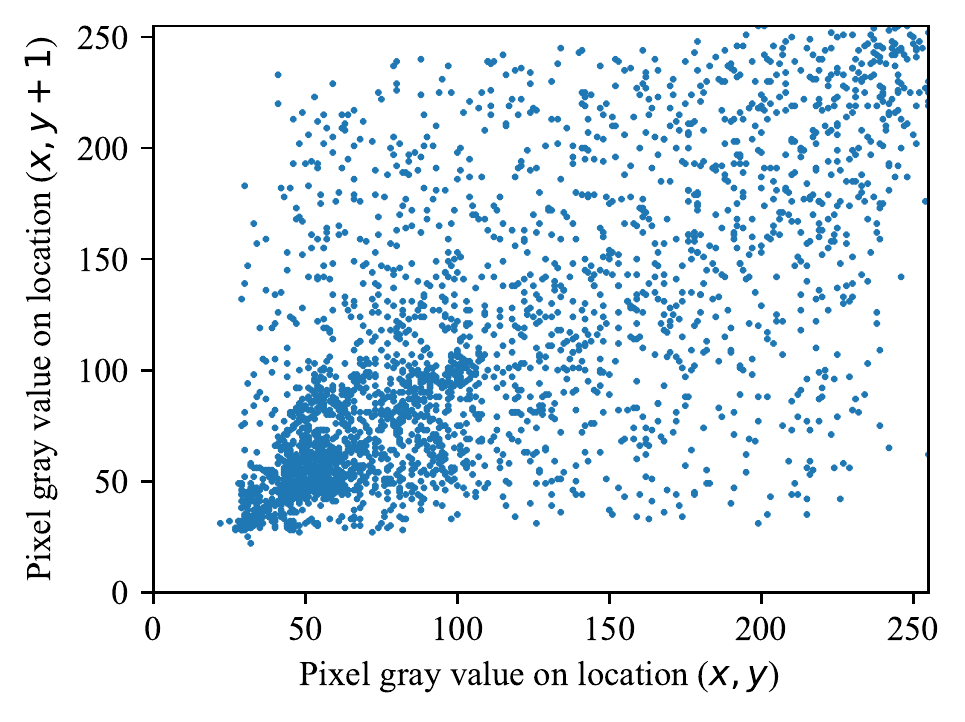}%
\label{fig:puppy_shf_v}}
\hfil
\subfloat{\includegraphics[width=0.33\linewidth]{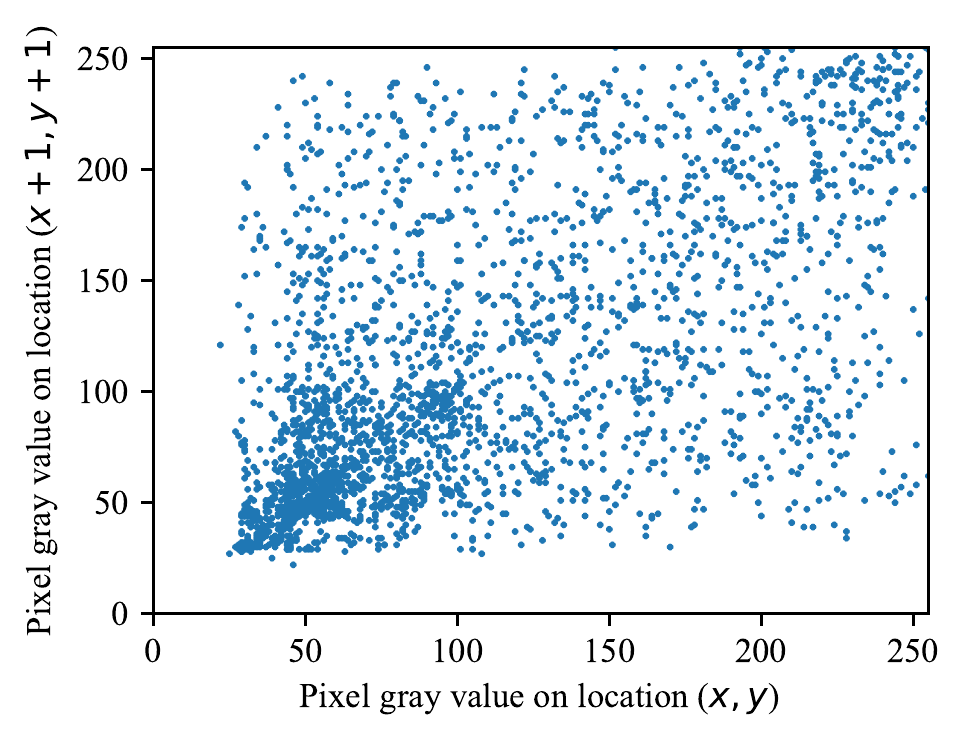}%
\label{fig:puppy_shf_d}}\\
\subfloat{\includegraphics[width=0.33\linewidth]{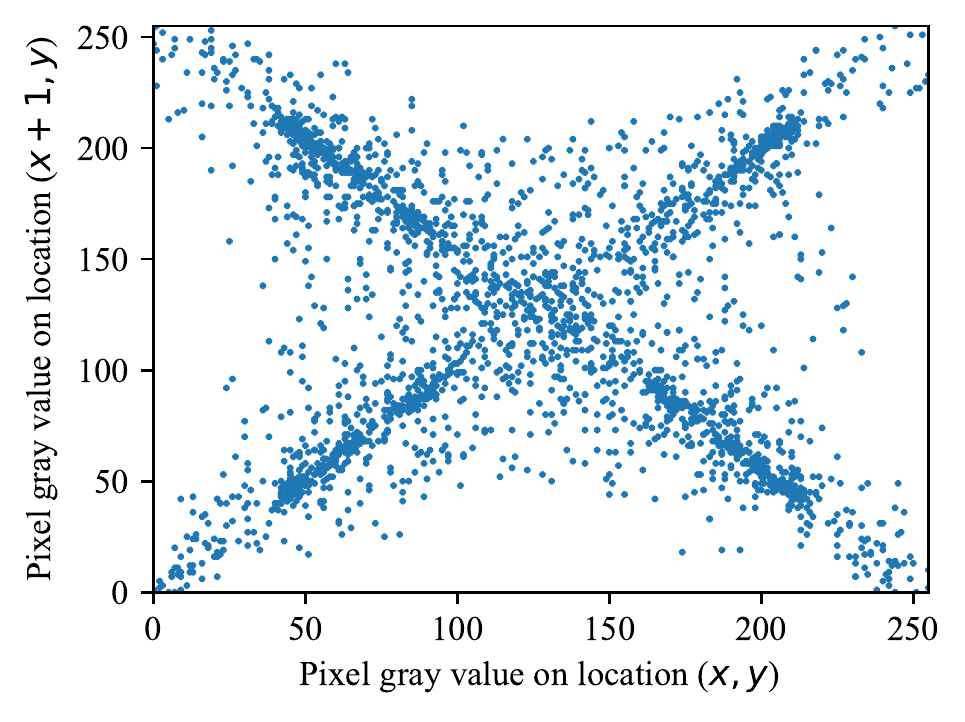}%
\label{fig:puppy_np_h}}
\hfil
\renewcommand{\thesubfigure}{c}
\subfloat[]{\includegraphics[width=0.33\linewidth]{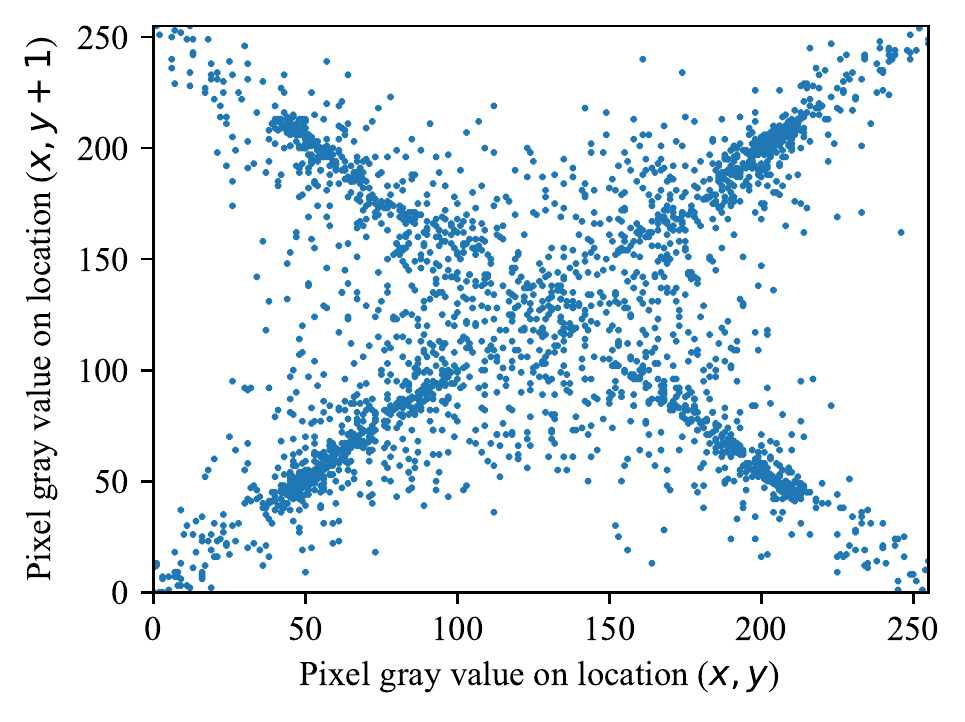}%
\label{fig:puppy_np_v}}
\hfil
\subfloat{\includegraphics[width=0.33\linewidth]{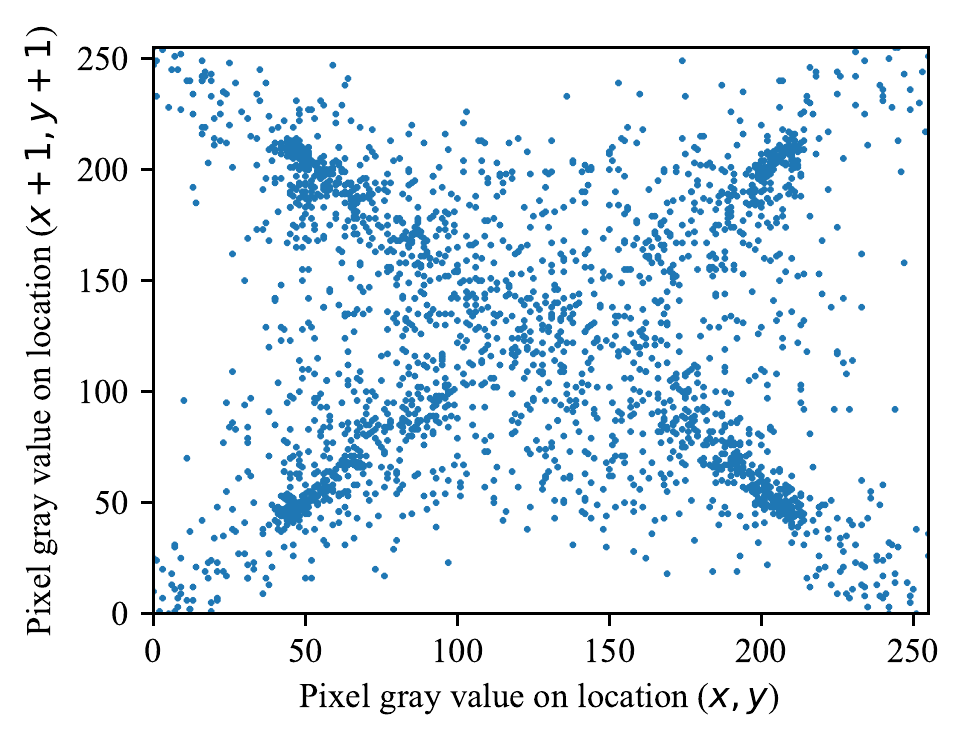}%
\label{fig:puppy_np_d}}\\
\subfloat{\includegraphics[width=0.33\linewidth]{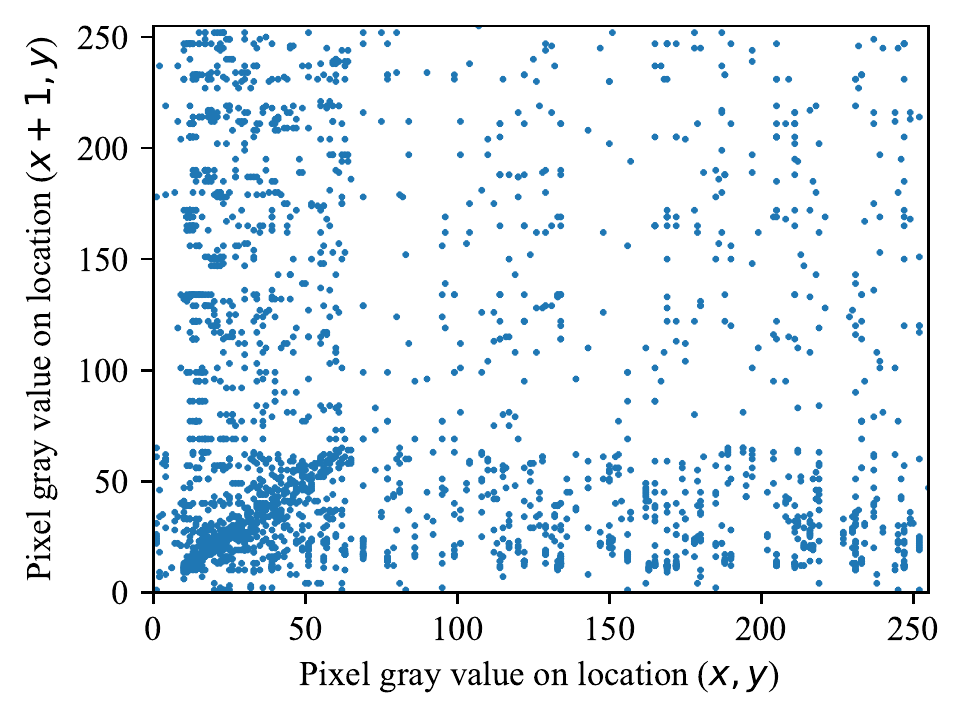}%
\label{fig:puppy_ffx_h}}
\hfil
\renewcommand{\thesubfigure}{d}
\subfloat[]{\includegraphics[width=0.33\linewidth]{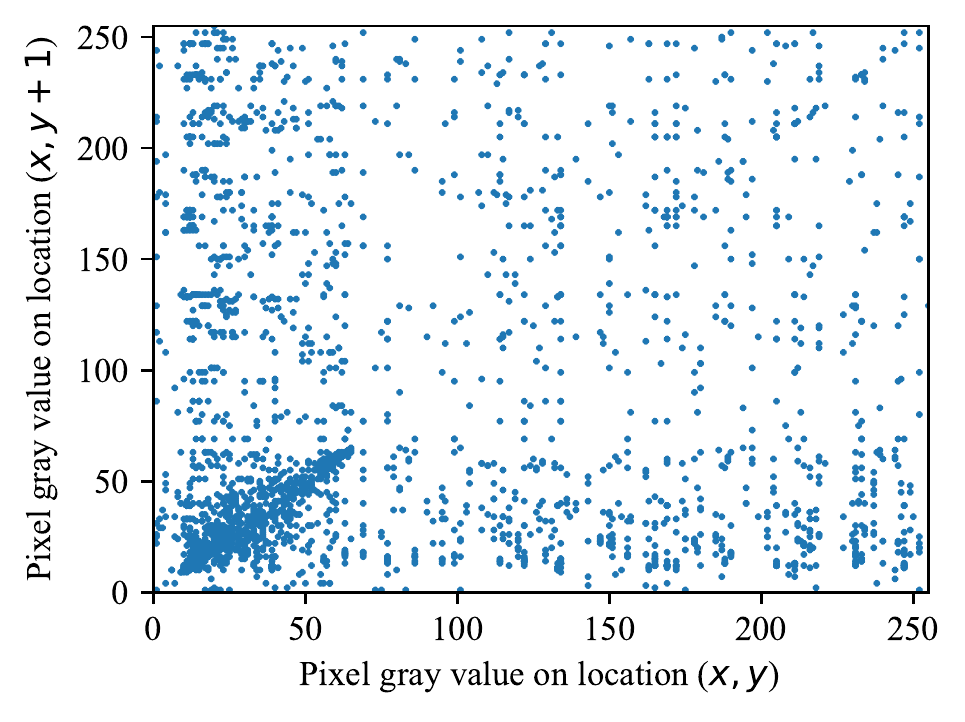}%
\label{fig:puppy_ffx_v}}
\hfil
\subfloat{\includegraphics[width=0.33\linewidth]{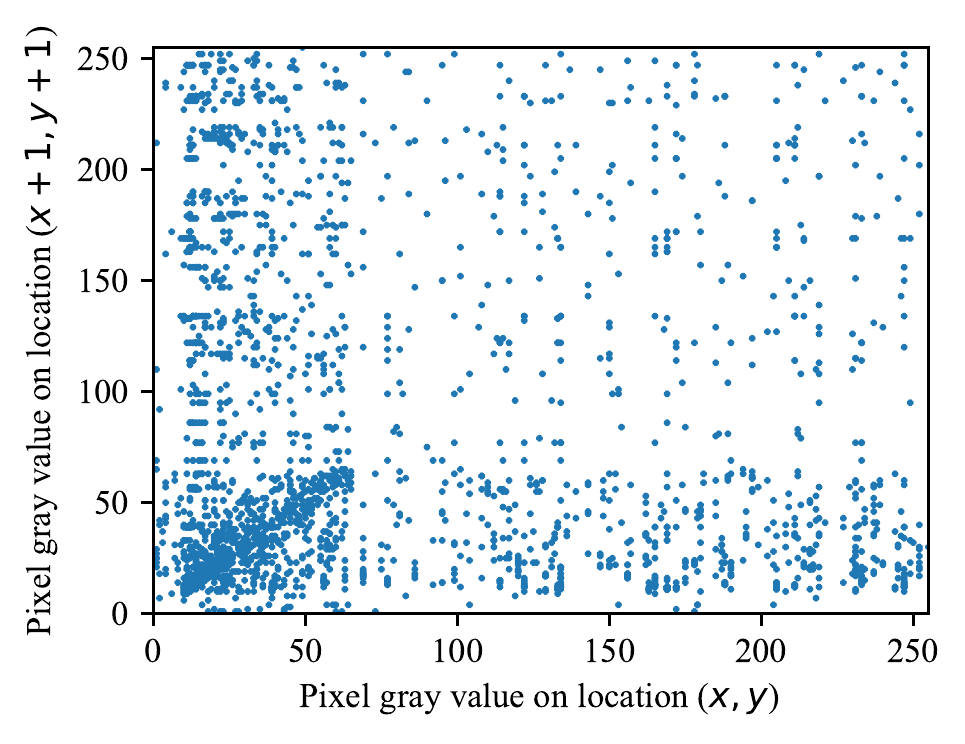}%
\label{fig:puppy_ffx_d}}\\
\caption{Horizontal, vertical and diagonal correlation test results for plain image and images transformed by SHF, NP and FFX ($M = 4$).\ (a) represents horizontal, vertical and diagonal correlation distribution of plain image, (b) represents that of image transformed by SHF, (c) represents that of image transformed by NP, and (d) represents that of image transformed by FFX.\label{fig:correlation}}
\end{figure*}

\section{Analysis and Discussion}
\label{sec:discussion}
\RA{In this section, we analyze the block-wise transformations utilized in the proposed model-protection method in terms of pixel correlation and key sensitivity. Then, we discuss possible key improvement, the selection of transformations, and the application range of the proposed scheme.}

\subsection{Image Correlation Analysis}
\label{sec:correlation}
\RB{To gain insights into the classification performance of block-wise transformed images, we carried out adjacent pixel correlation tests on a test image, ``dog,'' in the horizontal, vertical, and diagonal directions as shown in Fig.~\ref{fig:correlation}. From the figure, all transformations were confirmed to maintain some correlation between pixels differently. The pixel correlation distribution of the image transformed by SHF was similar to that of the plain image. For NP and FFX, the pixel correlation distributions were different from that of the plain image.} \RC{In addition, the pixel correlation of FFX was slightly weak due to the use of FFX, compared with the other ones, so this property might have caused a lower accuracy than those of the other transformations in Table~\ref{tab:results}. Accordingly, there is some correlation between pixels in the transformed images, so block-wise transformations can achieve a high classification accuracy.}

\subsection{Key Sensitivity Test}
\label{sec:sensitivity}
\RB{We carried out key sensitivity tests for models with $M = 4$ and $8$ on the CIFAR-10 and CIFAR-100 datasets. We define key sensitivity as the difference in accuracy between the correct key set and the modified key set (i.e., the correct key set with a small change), which is given by
\begin{equation}
  \text{Key Sensitivity} = \text{ACC} - \text{ACC}', \label{eqn:diff}
\end{equation}
where ACC is the classification accuracy with a correct key set, and ACC$'$ is that with a key set that has a small change from the correct key set.}

\RB{To make a small change, we swapped two random elements in the correct key for SHF, and one element in the correct key was flipped for NP and FFX (i.e., ``0'' to ``1'' and ``1'' to ``0''). Table~\ref{tab:sensitivity} shows the result of the key sensitivity tests, where the values in the table were averaged over $c \times M \times M$ times to cover changes in the different positions of the keys. From the results, we observed that low key sensitivity values reflected a higher accuracy for the incorrect keys, and high ones corresponded to a lower accuracy for the incorrect ones, as shown in Table~\ref{tab:results}.
The key sensitivity in the table gives some insights into the difference among transformations.
}

\robustify\bfseries
\sisetup{table-parse-only,detect-weight=true,detect-inline-weight=text,round-mode=places,round-precision=2}
\begin{table}[htbp]
\centering
\caption{Key sensitivity of various transformations with $M = 4$ and $8$\label{tab:sensitivity}}
\begin{tabular}{l|SS|SS}

  \toprule

& \multicolumn{2}{c|}{CIFAR-10} & \multicolumn{2}{c}{CIFAR-100}\\
  {Transformation} & {$M = 4$} & {$M = 8$} & {$M = 4$} & {$M = 8$}\\
\midrule
  {SHF} & 1.79 & 0.31 & 4.5 & 0.34\\
  {NP} & 9.68 & 0.11 & 18.9 & 1.97\\
  {FFX} & 5.17 & 0.29 & 10.43 & 0.52\\
  {SHF + NP} & 3.8 & 0.59 & 7.68 & 1.04\\
  {SHF + FFX} & 5.15 & 0.67 & 8.63 & 1.04\\
  {SHF + NP + FFX} & 5.65 & 0.72 & 10.13 & 1.06\\
  \bottomrule
\end{tabular}
\end{table}

\subsection{Discussion}
\RA{\textbf{Key Improvement:} When $M = 2$, the key space for the block-wise transformations is relatively small, so brute force attacks are possible. To improve the key space, there are two ways: (1) to use a larger block size such as $8 \times 8$, $8 \times 4$, etc. and (2) to use a combined transformation such as SHF + NP or SHF + FFX. Note that a value of $M$ affects not only the key space but also the classification accuracy and key sensitivity. Accordingly, users are requested to find a good trade-off among them.}

\RA{\textbf{Selection of Transformations:} Classification accuracy and model protection performance depend on the type of transformation and block size $M$. We recommend the following selection of transformations accordingly. When a higher classification accuracy is required, NP or a combined transformation such as SHF + NP or SHF + FFX with a small block size $M$ is recommended. When higher protection performance is preferred, a larger $M$ with SHF or a combined transformation is suitable.
}

\RA{\textbf{Application Range:} In this paper, the proposed model-protection method focuses on image classification tasks because the three encryption methods used in this paper are designed for image classification tasks. When these encryption methods are applied to other tasks such as image segmentation and image retrieval, the performance may drop compared with that of using plain images. Therefore, the proposed model protection is limited to image classification tasks, and novel image transformations are expected to be designed for applying other tasks.}

\section{Conclusion}
\label{sec:conclusion}
We proposed a model protection method that utilizes block-wise transformations with a secret key set to transform input images. Specifically, the transformation methods are pixel shuffling, negative/positive transformation, and format-preserving Feistel-based encryption. The performance accuracy of a protected model was closer to that of a non-protected model when the key set was correct, and it dropped drastically when an incorrect key set was given, suggesting that a protected model is not usable even when the model is stolen. The proposed method is also applicable to large datasets like the ImageNet dataset, which has never been tested by previous model-protection methods. Moreover, the proposed model-protection method does not introduce any overhead in both training and inference time. It is also robust against fine-tuning attacks in which the adversary has a small subset of a training dataset to adapt a new forged key set and key estimation attacks.

\bibliographystyle{unsrtnat}
\bibliography{refs}  






\end{document}